# Adaptive Intelligent Spider Robot

Rozita Teymourzadeh, *CEng, Member IEEE/IET*, Rahim Nakhli Mahal, Ng Keng Shen, and Kok Wai Chan
Faculty of Engineering, Technology & Built Environment
UCSI University
Kuala Lumpur, Malaysia
rozita@ucsiuniversity.edu.my

*Abstract* — This paper present the development of an adaptive intelligent spider robot. With the incorporation of sensors, the four-legged spider robot is able to monitor the environment wirelessly. A new auto-station adaptation National Instrument (NI) controller in application for intelligent robot is proposed. The research work will solve the weak adaptive ability of conventional existing robot. The proposed project proposes a self-adaptive smart spider robot that functions without human interfacing. The proposed design utilizes the control scheme with National Instruments (NI) LabVIEW to produce a smart portable system. Furthermore, the adaptive intelligent spider robot easily adapts to new situations when facing obstacles. The design is carried out using feed-back loop schematic LabVIEW interfacing with the smart controller and incorporation of GH-311 Ultrasonic Sensor to detect any obstacle in front of the robot, GH-312 Smoke Sensor which use to detect smoke in the particular area and LM35 Temperature sensor to get temperature for the surround. The proposed module was designed and implemented using movable wireless Router Module TP Link TL-MR3420 Router transceiver for device communication. The robot was tested and analyzed showing the system efficiency of above 95%, which is competent in robotic applications [1-4].

*Keywords* — *spider robot; adaptive controller; wireless sensor network; walking gait; National Instruments (NI) LabVIEW*

## I. Introduction

Nowadays there are many situations where human beings are unable to accomplish a certain task in the real life such as finding missing person in forest for more than 24 hours and exploring a cave with lack of oxygen. In order to accomplish these difficult tasks, human being will have to rely on mobile robots [5-9]. The significant challenge is to develop a self-adaptive system for the mobile robots in adapting towards uncertain changes in the environment. Researchers today begin to focus on the novel design of self-adaptive robotic system which include trajectory tracking [10].

In 2011, Roy, Singh, and Pratihar [11] evaluated the optimal feet forces and joint torques in the real-time system for controlling of six-legged robot. Their research is focused on obtaining an optimum point in the distributions of feet forces and values of joint torques of a six-legged robot on-line. The minimization of norm of both feet forces and joint torques are simulated in their research. Another research conducted in the same year, Roy and Pratihar [12] found out that different duty factors will lead to various energy consumption. The duty factor can vary between 1/3 and 2/3 while energy consumption will change in the range of 2% and 35%. In the following year, both Roy and Pratihar [13] conducted research on legged robot and simulated the method of achieving maximum stability with the least energy consumption gaits. Boscariol, Henry, and Menon [14], in 2013, investigated on solving the inappropriate redistribution of forces in the preloading of the legs in climbing robots which will lead to irreparable detachment of the robots from vertical surfaces. They concluded that it is possible to save 35% of the total cost of the robot if the designed gait is efficient.

In this research work, National Instrument (NI) protocol has been selected to communicate with the proposed smart robotic controller for processor and interfacing. The proposed intelligent system collects information about the surroundings especially in accessible areas and helps the spider robot to decide the best path direction to be taken. NI embedded field-programmable gate array (FPGA) board is selected to incorporate with spider robot for the purpose of high performance efficiency (floating point architecture) as well as being user-friendly and compatibility with LabVIEW graphic interfacing. The sequence of leg movements is predefined in this research for continuous walking. With the incorporation of multiple sensors on the body structure, this proposed four-legged adaptive intelligent spider robot is suitable to be implemented in uncertain terrains.

## II. Stage Realization

NI Smart Control System and Automation System for Spider Robot is developed to execute certain challenging tasks that human beings have difficulty to accomplish. The proposed robotic system is divided into three subsystems, which include body structure of the robot, sensor and wireless data acquisition, and walking control algorithm.

As shown in Fig. 1, the structure of the spider robot is made of aluminium due to the strength of aluminium in withstanding the pressure and tensile stress on its surface. Standoff screws are used to secure the position of the NI board on the chassis. Multiple sensors are implemented in the robot body [15-18]. The body of the robot must be strong enough to support the weight of the NI board in dynamic situations. Hence, the consideration taken during the design stage for the main body of the robot include tensile and compression stress along the axis of the movement of the moving parts of the robot caused by the payload carried.

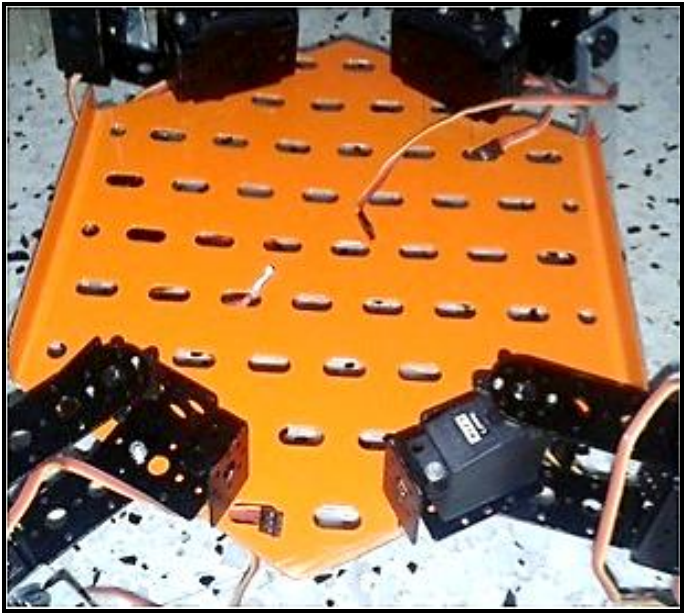

Fig. 1. Aluminium structure of proposed robot.

While a servo bracket connected to the joint of the leg, each individual leg of the robot has three degree of freedom (DOF). One of the degree of freedom is along the $z$-axis and serves as the shoulder and the rotation of the shoulder will control the other two joints. Shoulder will be master and the position of it will force the other two to take their position. NI board controls the movement of the joints.

Fig. 2 shows the robot legs with 3 joints in the early stage of the experiment. Each of the joint uses a servo bracket to joint to each other. A holder mounts the servo motor and is able to connect each other with servo bracket U joint.

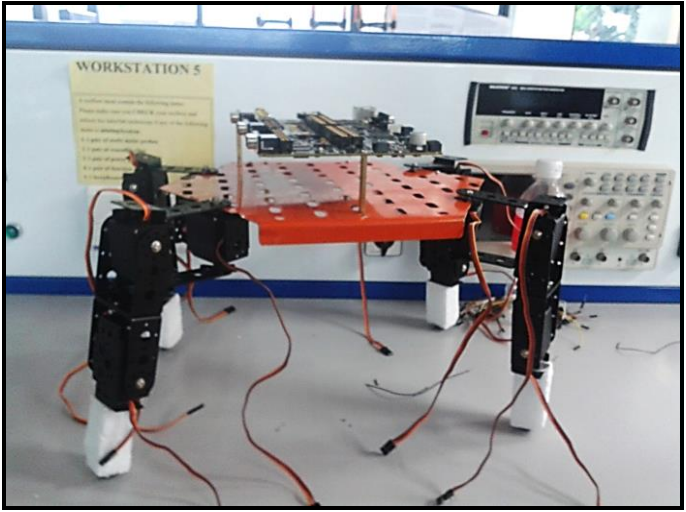

Fig. 2. Joints using U-brackets for the robot legs.

The overall design of the self-adaptive system is carried out using feedback loop schematic NI LabVIEW interfacing with smart controller and corporation of GH-311 ultrasonic sensor, smoke sensor, and LM35 temperature sensor. The GH-311 ultrasonic sensor provides a precise and non-contact distance measurement. Whenever there is an obstacle as far as 30cm away from the robot, the sensor will transmit a high signal to the NI board to indicate the location of obstacle and execute the appropriate walking algorithm to avoid the obstacle. The GH-312 sensor is utilized in this robot structure to detect smoke, liquefied gas, butane and propane, alcohol, and hydrogen.

The homogeneous transformation matrix describing the relative translation and rotation between the $i^{th}$ and $(i-1)^{th}$ coordinate system is represented in (1) [11].

$$^{i-1}T_i = \begin{bmatrix} \cos q_i & -\sin q_i \cos \alpha_i & \sin q_i \sin \alpha_i & a_i \cos q_i \\ \sin q_i & \cos q_i \cos \alpha_i & -\cos q_i \sin \alpha_i & a_i \sin q_i \\ 0 & \sin \alpha_i & \cos \alpha_i & d_i \\ 0 & 0 & 0 & 1 \end{bmatrix} \quad (1)$$

where $q_i$ denotes the joint angle of $x_i$ axis relative to $x_{i-1}$ axis with defined according to right-hand (RH) rule about $z_{i-1}$ axis, while $\alpha_i$ indicates the offset angle of $z_i$ axis relative to $z_{i-1}$ axis measured about the $x_i$ axis using RH rule. The coefficient $d_i$ represents the distance from the origin of the $i-1$ axes to the intersection of the $z_{i-1}$ axis with the $x_i$ axis and measured along the $z_{i-1}$ axis. The parameter $a_i$ designates the minimum distance between $z_{i-1}$ and $z_i$. Solving parametrically for the three joints using the Denavit-Hartenberg (D-H) parameters will lead to the definition of the 3-degree of freedom (DOF) for the robotic legs in (2), where the position of the robot's leg is represented by $p_x$, $p_y$, and $p_z$. [13].

$$^0T_3 = \begin{bmatrix} \cos q_1 \cos(q_2+q_3) & -\cos q_1 \sin(q_2+q_3) & \sin q_1 & p_x \\ \sin q_1 \cos(q_2+q_3) & -\sin q_1 \sin(q_2+q_3) & -\cos q_1 & p_y \\ \sin(q_2+q_3) & \cos(q_2+q_3) & 0 & p_z \\ 0 & 0 & 0 & 1 \end{bmatrix} \quad (2)$$

The joint angles for Link 1 ($L_1$), Link 2 ($L_2$), and Link 3 ($L_3$) of the self-adapting spider robot's leg can be determined from the calculation based on (3), (4) and (5) [13].

$$q_1 = a\tan2(p_y, p_x) \quad (3)$$

$$q_2 = a\tan2(c, \pm\sqrt{a^2 + b^2 - c^2}) - a\tan2(a, b) \quad (4)$$

$$q_3 = \cos^{-1}\left(\frac{c}{2L_2 L_3}\right) \quad (5)$$

where,

$a = 2L_2(\sqrt{p_x^2 + p_y^2} - L_1);$

$b = 2p_z L_2;$

$c = [(\sqrt{p_x^2 + p_y^2} - L_1)^2 + p_z^2 + L_2^2 - L_3^2]$

### III. IMPLEMENTATION

As portrayed in Fig. 3, the spider robot is programmed in NI LabVIEW [19, 20] with the aim of moving forward-backward and avoiding the obstacles on the robot's path. NI

sbRIO-9642XT board is pre-programmed and waits for user to input a certain task to be executed. The NI board collects feedback data from sensors embedded on the robot body and processes the data; hence, executing the assigned task. For example, if the ultrasonic sensor senses an object blocking the robot's movement path, the sensor will transmit the feedback data to the NI board. The NI board will then process the data accordingly and evaluates the data giving an appropriate result such as refinement for the position of the robot to avoid the obstacle that leads to self-localization. The robot will then start moving forward again until the robot reaches destination or encountering another obstacle on the movement path.

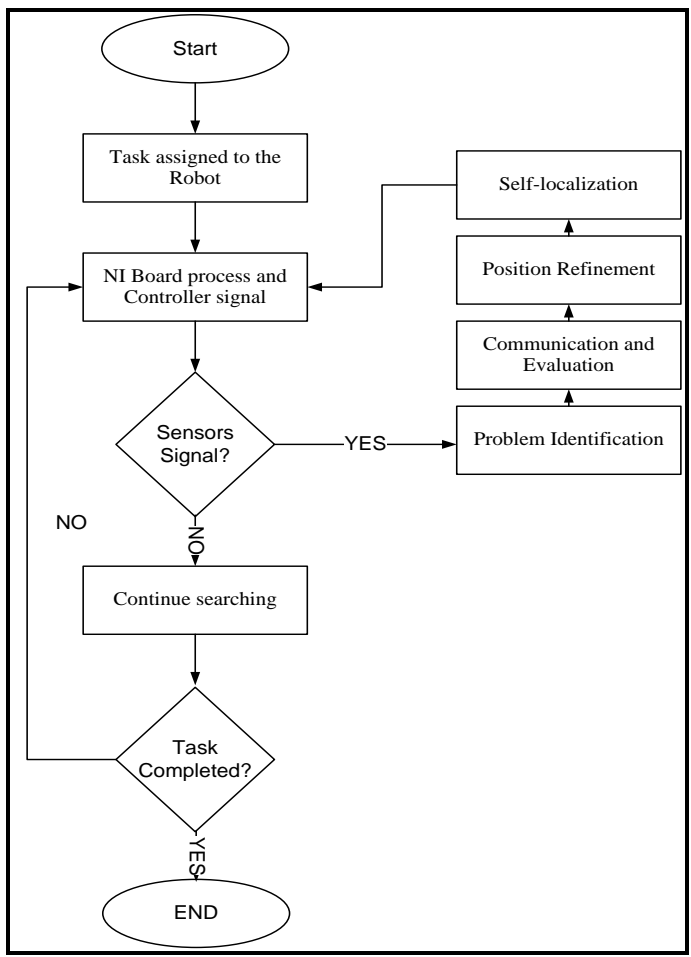

Fig. 3. Flow chart of the robotic system.

The entire system algorithm is presented in Fig. 4. There are three Digital I/O input ports in the program, which include ultrasonic sensor, smoke sensor, and temperature sensor. In the LabVIEW block diagram, there are four types of walking algorithms ready to be executed respectively whenever the ultrasonic sensor detects any obstacle in front of the robot. When there is no obstacle on the robot's path, three of the walking algorithms will not be executed. The only walking algorithm executed is the algorithm for the robot moving forward. If the ultrasonic sensor detects an obstacle in front of the robot, the robot will terminate the walking forward algorithm and then activate either walking left, right or backward walking algorithm. The feedback data from the sensor will be executed accordingly; hence, calling the data that has been generated in memory block to the pulse-width-modulated (PWM) generator for each of the servo motors.

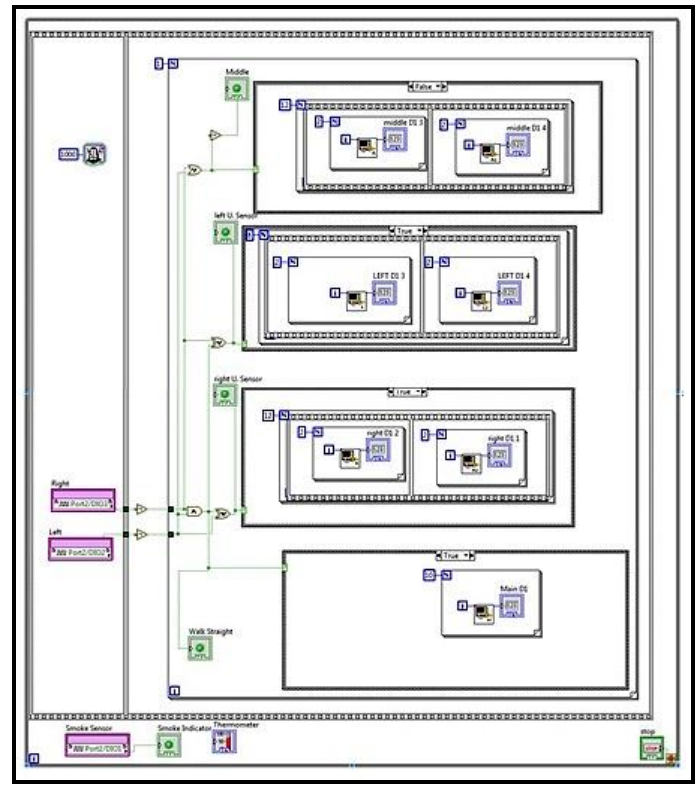

Fig. 4. Robot system algorithm in LabVIEW.

When the forward walking algorithm is executed, the robot leg moves step-by-step. With the four legs taking their respective position, the robot will then push the body forward. This process will repeat to create a forward movement for the robot. Fig. 5 presents the computer aided design (CAD) simulation of the robot's leg movement at different angles to generate a move when executing the forward walking algorithm.

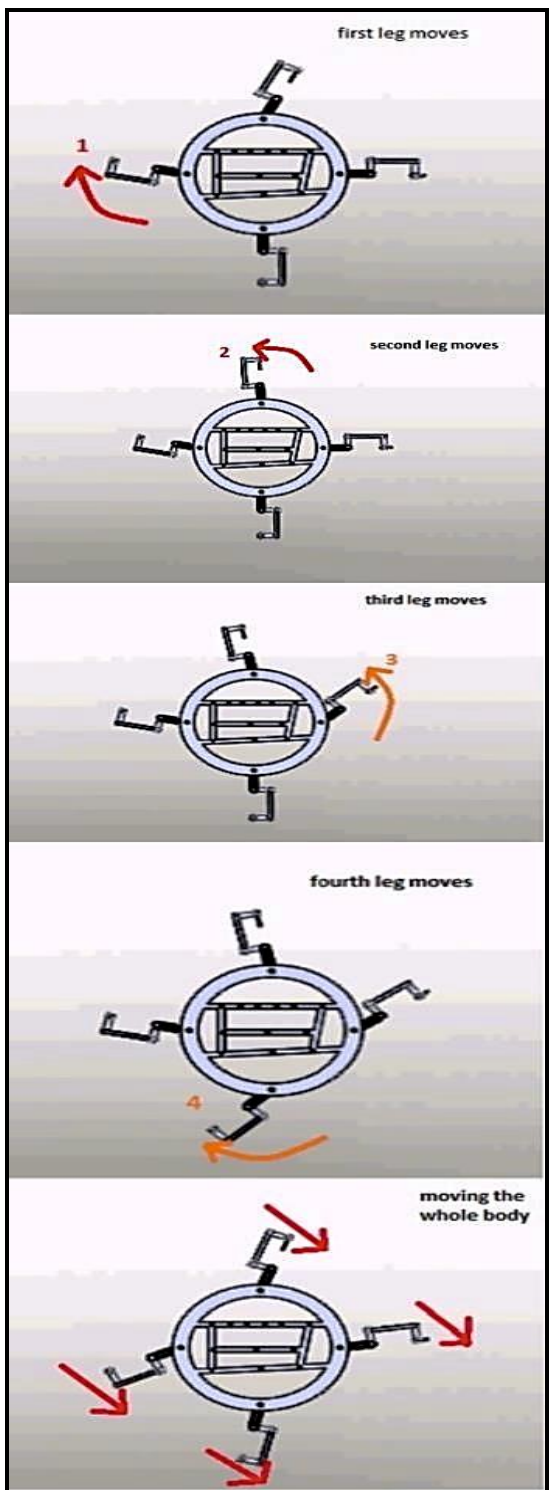

Fig. 5. Simulation of robot movement in CAD.

The temperature sensor and smoke sensor will work independently without affecting the walking algorithm of the spider robot. The data obtained from both the temperature sensor and smoke sensor will be displayed on the NI LabVIEW front panel. This wireless monitoring system interface on laptop is illustrated in Fig. 6. On the monitoring system interface, green LED indicators represent the respective the direction path taken by the spider robot when walking algorithm is executed. In the event that the spider robot detects smoke, the red LED indicator will light up to signal potential threat. The surrounding temperature is being monitored and updated every 0.5 seconds with the indicator thermometer on the LabVIEW monitoring system.

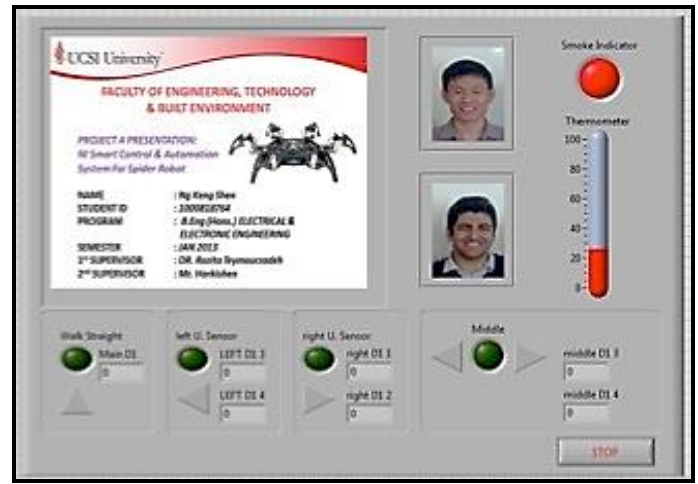

Fig. 6. LabVIEW monitoring system.

## IV. RESULTS AND DISCUSSIONS

The angles for feasible positions of each leg have been generated for calculation of the robot's walking gait. For forward walking motion, the two front legs of the robot will need to utilize the front ¾ part of their reachable range whereas the two rear legs of the robot will use the rear ¾ part of their reachable range. Fig. 7, Fig. 8, Fig. 9, and Fig. 10 respectively shows the change in the angles of the robot's leg joints with respect to the robot's walking sequence. For example, the first joint of the robot's front right leg is represented by L11 in Fig. 7. Based on the walking gait, the leg movement sequence of Leg1 →Leg3 →Leg4 →Leg2 will be repeated for the robot to continue walking forward.

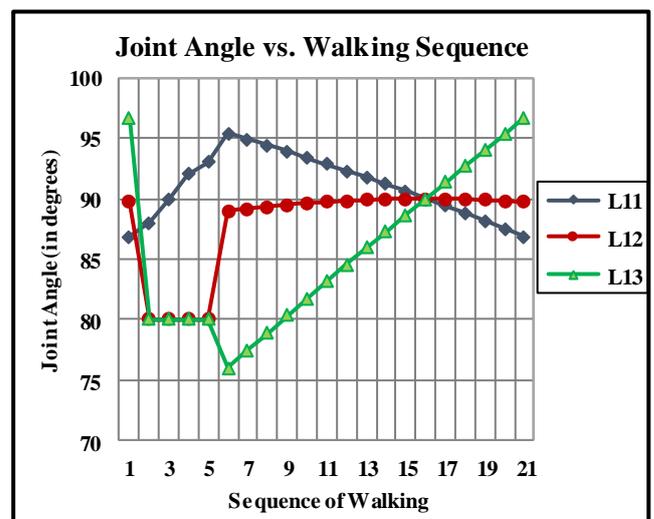

Fig. 7. Changes in angles of respective joint with respect to sequence of walking for front right leg.

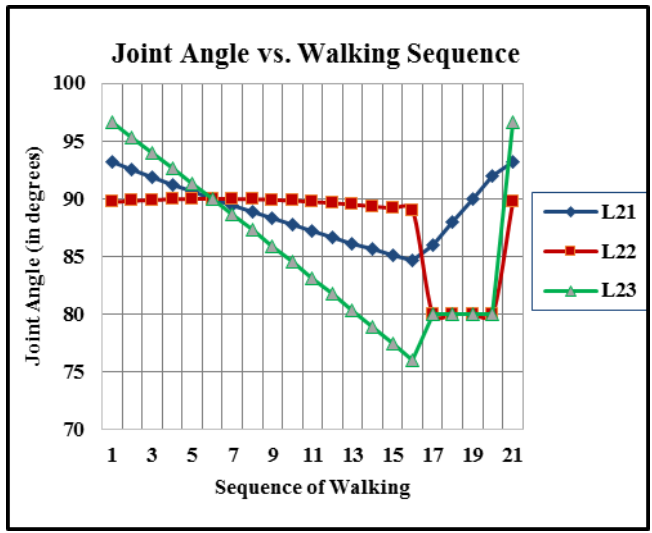

Fig. 8. Changes in angles of respective joint with respect to sequence of walking for front left leg.

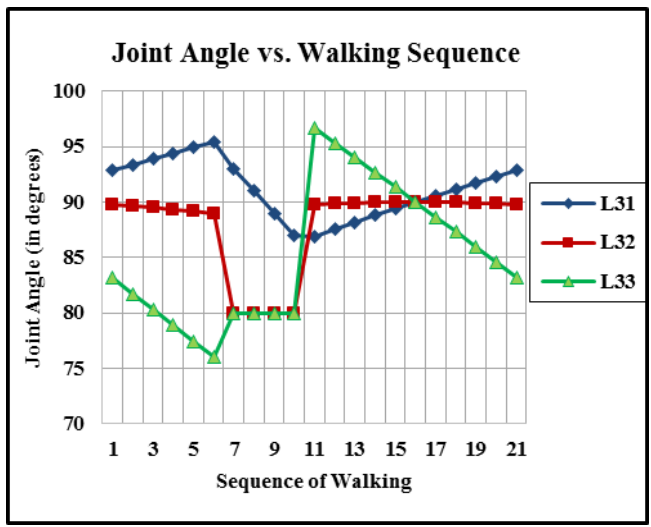

Fig .9. Changes in angles of respective joint with respect to sequence of walking for back left leg.

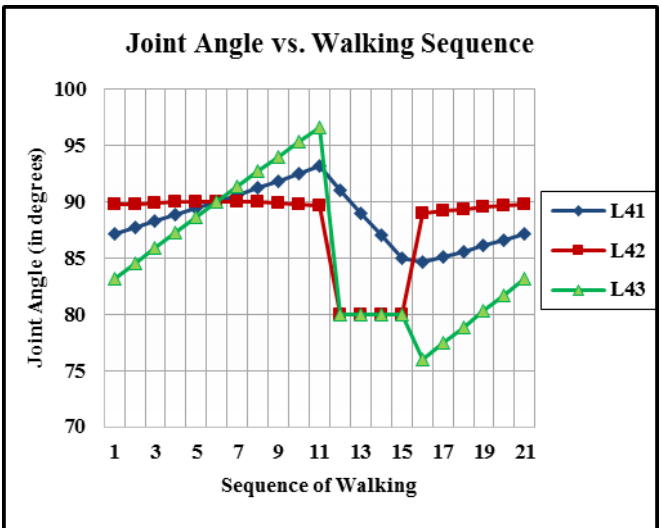

Fig. 10. Changes in angles of respective joint with respect to sequence of walking for back right leg.

The sensors embedded onto the spider robots have been tested to verify the measurement accuracy. Fig. 11 illustrates the statistical result of the accuracy in temperature measurement using the temperature sensor. As compared to the real temperature of the surrounding area, the error rate in temperature measurement using LM35 temperature sensor is less than 5%.

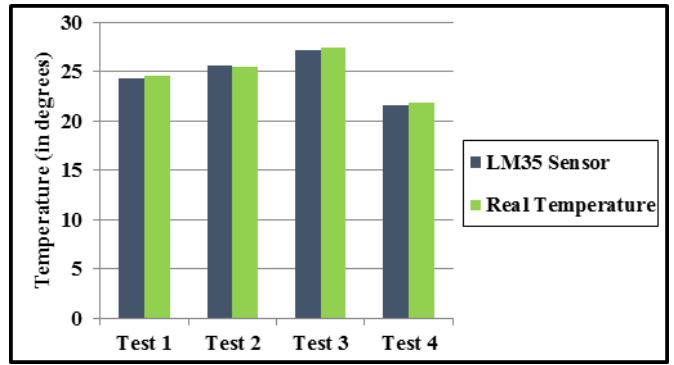

Fig. 11. Statistical result for accuracy of temperature measurement.

As shown in Fig. 12, the performance of both ultrasonic sensors is respectively competent in ultrasonic non-contact detection readings taken in the tests conducted, with an average performance of ≥ 95%. The gas sensor embedded onto the adaptive intelligent spider robot, as illustrated in Fig. 13 has a system efficiency of 100% in gas sensor detection readings during the system testing.

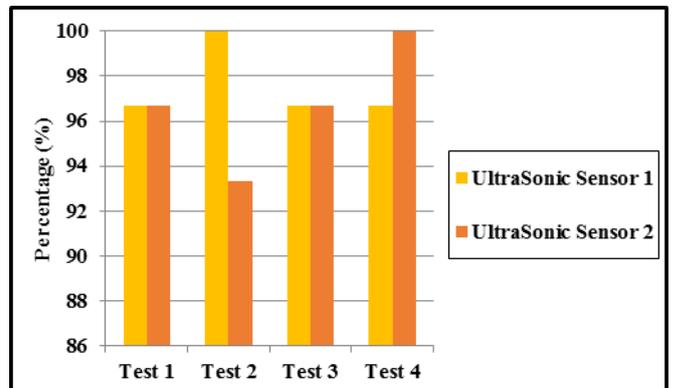

Fig. 12. Ultrasonic sensor performance during system testing.

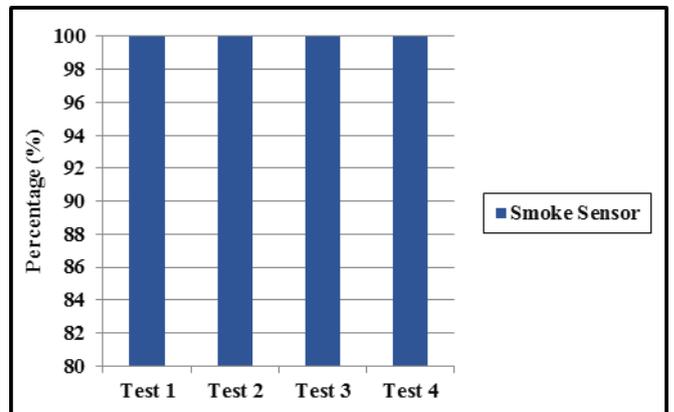

Fig. 13. Smoke sensor performance during system testing.

## V. Conclusion

An intelligent self-adaptive spider robot equipped with NI LabVIEW has been designed and implemented. With the integration of the designed NI Smart Control & Automation System, the spider robot system is able to monitor the surrounding, analyse the scenario, plan, and execute the appropriate action. The statistical experiments have been performed and proven that the spider robot's LabVIEW monitoring system incorporating sensors have system efficiency of ≥95%, which is competent in robotic applications.